%% file: main.tex
\documentclass{article}
\usepackage{academicons}

\usepackage{authblk}
\usepackage[utf8]{inputenc}
\usepackage{main}
\usepackage{microtype}
\usepackage{subcaption}
\usepackage{graphicx}
\usepackage{times}
\usepackage{latexsym}
\usepackage{amsmath}
\usepackage{float}
\usepackage{footnote}
\usepackage{enumitem}
\usepackage{bm}
\usepackage{arydshln}
\usepackage{booktabs}
\usepackage{array}     
\usepackage{multicol}
\usepackage{multirow}
\usepackage{color}
\usepackage{xcolor}     
\usepackage{colortbl}
\usepackage{bbding}
\usepackage{makecell}
\usepackage{mathtools}
\usepackage{imakeidx}
\usepackage{longtable}
\usepackage{wrapfig}
\usepackage{algorithmic}
\usepackage{rotating}
\makeindex
\usepackage{arydshln}
\usepackage{lipsum}
\usepackage{natbib}
\usepackage[toc]{multitoc}
\usepackage[edges]{forest}
\usepackage[normalem]{ulem}
\definecolor{mydarkblue}{rgb}{0,0.08,0.45}
\usepackage[colorlinks=true,linkcolor=mydarkblue,citecolor=mydarkblue,filecolor=mydarkblue,urlcolor=mydarkblue]{hyperref}
\usepackage{CJKutf8}
\usepackage{awesomebox} 
\usepackage{bbding}
\usepackage[most]{tcolorbox}
\usepackage{booktabs}
\usepackage{geometry}
\definecolor{wkblue}{rgb}{0.2, 0.3, 0.6}
\definecolor{meta-color}{rgb}{0.5, 0.5, 0.5}
\usepackage{amsmath}
\usepackage{enumitem}
\usepackage{lscape} 
\usepackage{booktabs}
\usepackage{algorithm}   
\usepackage{setspace}

\usepackage{algorithmic}
\usepackage{tabularx,booktabs}
\usepackage{makecell}

\usepackage{amssymb}
\usepackage{amsfonts}

\usepackage{booktabs} 
\usepackage{geometry} 

\usepackage{multirow}

\usepackage[tikz]{bclogo}
\usepackage[framemethod=tikz]{mdframed}
\definecolor{bgblue}{RGB}{245,243,253}
\definecolor{ttblue}{RGB}{91,194,224}

\usepackage{pgfplots}
\usepackage{pgfplotstable}

\usepackage{wrapfig}
\usepackage{graphicx}

\mdfdefinestyle{mystyle}{%
  rightline=true,
  innerleftmargin=10,
  innerrightmargin=10,
  outerlinewidth=3pt,
  topline=false,
  rightline=true,
  bottomline=false,
  skipabove=\topsep,
  skipbelow=\topsep
}

\newtcolorbox{myboxi}[1][]{
  breakable,
  title=#1,
  colback=red!5,
  colbacktitle=red!5,
  coltitle=black,
  fonttitle=\bfseries,
  bottomrule=0pt,
  toprule=0pt,
  leftrule=2pt,
  rightrule=2pt,
  titlerule=0pt,
  arc=0pt,
  outer arc=0pt,
  colframe=red,
}

\newtcolorbox{myboxnote}[1][]{
  breakable,
  title=#1,
  colback=orange!0,
  colbacktitle=orange!0,
  coltitle=black,
  fonttitle=\bfseries,
  bottomrule=0pt,
  toprule=0pt,
  leftrule=2pt,
  rightrule=2pt,
  titlerule=0pt,
  arc=0pt,
  outer arc=0pt,
  colframe=orange,
}

\newtcolorbox{myboxii}[1][]{
  breakable,
  freelance,
  title=#1,
  colback=white,
  colbacktitle=white,
  coltitle=black,
  fonttitle=\bfseries,
  bottomrule=0pt,
  boxrule=0pt,
  colframe=white,
  overlay unbroken and first={
  \draw[red!75!black,line width=3pt]
    ([xshift=5pt]frame.north west) -- 
    (frame.north west) -- 
    (frame.south west);
  \draw[red!75!black,line width=3pt]
    ([xshift=-5pt]frame.north east) -- 
    (frame.north east) -- 
    (frame.south east);
  },
  overlay unbroken app={
  \draw[red!75!black,line width=3pt,line cap=rect]
    (frame.south west) -- 
    ([xshift=5pt]frame.south west);
  \draw[red!75!black,line width=3pt,line cap=rect]
    (frame.south east) -- 
    ([xshift=-5pt]frame.south east);
  },
  overlay middle and last={
  \draw[red!75!black,line width=3pt]
    (frame.north west) -- 
    (frame.south west);
  \draw[red!75!black,line width=3pt]
    (frame.north east) -- 
    (frame.south east);
  },
  overlay last app={
  \draw[red!75!black,line width=3pt,line cap=rect]
    (frame.south west) --
    ([xshift=5pt]frame.south west);
  \draw[red!75!black,line width=3pt,line cap=rect]
    (frame.south east) --
    ([xshift=-5pt]frame.south east);
  },
}

\usepackage{fontawesome5}
\usepackage{fancyhdr} 
\usepackage{blindtext} 
\usepackage{makecell}

\DeclareCaptionFont{black}{\color{black}}

\definecolor{myblue}{rgb}{0.9, 0.1, 0.94}
\definecolor{mygreen}{rgb}{0.64, 0.56, 0.88}
\definecolor{myyellow}{rgb}{0.68, 0.6, 0.1}
\definecolor{fancygreen}{rgb}{0.33, 0.68, 0.20}
\definecolor{salmon}{rgb}{0.94, 0.52, 0.49}
\definecolor{tablegreen}{rgb}{0.82, 0.94, 0.75}
\definecolor{tableblue}{rgb}{0.81, 0.90, 0.94}
\definecolor{tablered}{rgb}{0.97, 0.85, 0.85}
\definecolor{tableorange}{rgb}{0.96, 0.85, 0.81}

\newenvironment{itemize*}%
 {\leftmargini=10pt\begin{itemize}%
  \setlength{\itemsep}{0pt}%
  \setlength{\parskip}{0pt}%
  }%
 {\end{itemize}}
\newenvironment{enumerate*}%
 {\begin{enumerate}%
  \setlength{\itemsep}{0pt}%
  \setlength{\parskip}{0pt}}%
 {\end{enumerate}}

\usepackage{xcolor}
\usepackage{listings}  

\newcommand\JSONnumbervaluestyle{\color{blue}}
\newcommand\JSONstringvaluestyle{\color{red}}

\newif\ifcolonfoundonthisline

\makeatletter

\lstdefinestyle{json}
{
  showstringspaces    = false,
  keywords            = {false,true},
  alsoletter          = 0123456789.,
  morestring          = [s]{"}{"},
  stringstyle         = \ifcolonfoundonthisline\JSONstringvaluestyle\fi,
  MoreSelectCharTable =%
    \lst@DefSaveDef{`:}\colon@json{\processColon@json},
  basicstyle          = \ttfamily,
  keywordstyle        = \ttfamily\bfseries,
}

\newcommand\processColon@json{%
  \colon@json%
  \ifnum\lst@mode=\lst@Pmode%
    \global\colonfoundonthislinetrue%
  \fi
}

\lst@AddToHook{Output}{%
  \ifcolonfoundonthisline%
    \ifnum\lst@mode=\lst@Pmode%
      \def\lst@thestyle{\JSONnumbervaluestyle}%
    \fi
  \fi
  \lsthk@DetectKeywords%
}

\lst@AddToHook{EOL}%
  {\global\colonfoundonthislinefalse}

\makeatother

\usepackage{etoolbox}
\usepackage{natbib}
\usepackage{url}
\newcounter{bibcount}
\makeatletter
\patchcmd{\@lbibitem}{\item[}{\item[\hfil\stepcounter{bibcount}{[\thebibcount]}}{}{}
\renewcommand\NAT@bibsetup%
  [1]{\setlength{\leftmargin}{\bibhang}\setlength{\itemindent}{-\parindent}%
      \setlength{\itemsep}{\bibsep}\setlength{\parsep}{\z@}}
\makeatother

\definecolor{mybrown}{RGB}{128,64,0}

\definecolor{titlecolor}{HTML}{4c9cff}
\definecolor{coolblue3}{rgb}{0.91, 0.94, 0.98}

\input{preamble}

\begin{document}



\title{\textsc{AgencyBench}: Benchmarking the Frontiers of Autonomous Agents in 1M-Token Real-World Contexts}

\author[1,2,4]{Keyu Li}
\author[2,4]{Junhao Shi}
\author[3,4]{Yang Xiao}
\author[1,2,4]{Mohan Jiang}
\author[1,4]{Jie Sun}
\author[2,4]{Yunze Wu}
\author[1,4]{Dayuan Fu}  
\author[1,2,4]{Shijie~Xia}
\author[2,4]{Xiaojie Cai}
\author[2,4]{Tianze Xu} 
\author[2,4]{Weiye Si} 
\author[3]{Wenjie Li}
\author[1,2,4]{Dequan Wang\textsuperscript{†}}
\author[1,2,4]{Pengfei Liu\textsuperscript{†}}
\affil{SII \quad \textsuperscript{2}SJTU \quad \textsuperscript{3}PolyU \quad \textsuperscript{4}GAIR}

\maketitle
\pagestyle{fancy}
\thispagestyle{fancy}
\fancyhead{}

\lhead{
  \raisebox{-0.2cm}{\includegraphics[height=0.85cm]{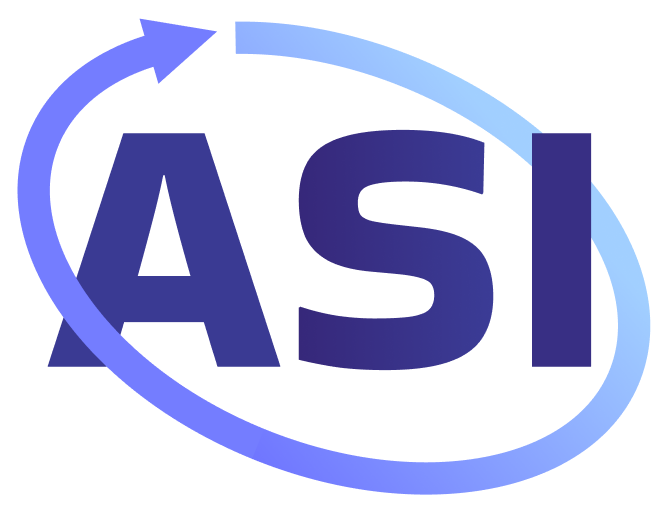}}
}
\rhead{%
  \raisebox{-0.2cm}{\includegraphics[height=0.7cm]{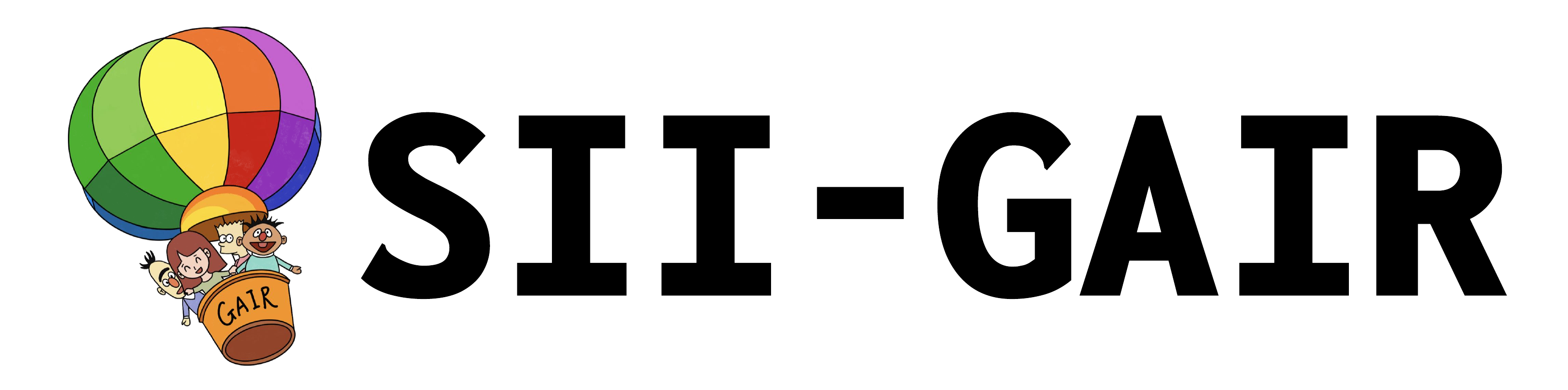}}%
}

\renewcommand{\headrulewidth}{0pt}
\setlength{\headsep}{-0.8mm} 


\renewcommand{\thefootnote}{}
\footnotetext{† Corresponding authors.}
\vspace{-20pt}

{\centering
\href{https://github.com/sii-research/GAIR}{\raisebox{-.15em}{\includegraphics[height=1em]{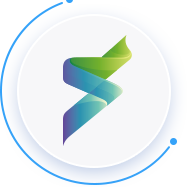}}\ SII Open Source:}
\quad\href{https://agencybench.opensii.ai}{\raisebox{-.15em}{\includegraphics[height=1em]{assets/sii.png}}\ AgencyBench}
\quad \href{https://github.com/GAIR-NLP/AgencyBench}{\textcolor{black}\faGithub\ Code}
\quad\href{https://huggingface.co/datasets/GAIR/AgencyBench}{\raisebox{-.15em}{\includegraphics[height=1em]{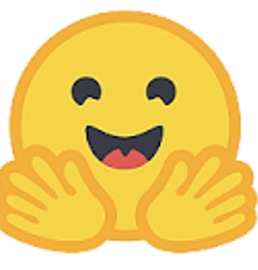}}\ AgencyBench}
\par}


\input{sec/abstract}

\input{fig/fig_tex/teaser}

\newpage 

\lhead{\rightmark}

\renewcommand{\headrulewidth}{0.7pt} 
\setlength{\headsep}{5mm}

\input{sec/introduction}

\input{sec/related}

\input{sec/method}

\input{sec/exp}

\input{sec/conclusion}

\section*{Limitations}
\paragraph{Model Selection Coverage}
While \textsc{AgencyBench} evaluates a diverse set of representative proprietary and open-source models, the landscape of Large Language Models evolves rapidly. Due to constraints on computational resources and budget, our evaluation cannot exhaustively cover every emerging model variant, intermediate checkpoint, or specialized fine-tune. Consequently, our findings reflect a snapshot of the current state-of-the-art and may not capture the full performance spectrum of models released subsequent to our experiments.

\paragraph{Domain Specificity}
Our benchmark focuses on high-complexity tasks within digital environments, such as game development, software engineering, web research, etc. While these scenarios encompass 32 real-world situations, they are confined to software-based agents operating within a computer interface. The current framework does not extend to embodied agents or tasks requiring physical world interaction (e.g., robotics), leaving the evaluation of such multimodal physical agency for future research.

\section*{Ethical Considerations}
\paragraph{Human Subjects and Compensation}
The construction of \textsc{AgencyBench} involved surveys and data validation by human experts, specifically computer science researchers and developers. We strictly adhere to the ACL Code of Ethics regarding human participant research. All contributors were fully informed of the project's scope, their data was anonymized to protect privacy, and they were compensated at a rate significantly exceeding the local hourly minimum wage to ensure fair and ethical treatment.

\paragraph{Safety and Usage}
Given that our benchmark involves agents generating executable code and performing shell operations, there are inherent risks associated with autonomous execution. To mitigate this, all evaluations are strictly confined within isolated Docker containers (Remote Sandbox) with controlled network access, preventing any potential harm to host systems. We will release this benchmark to foster the development of reliable and safe autonomous agents and explicitly oppose the application of these capabilities for malicious purposes, such as automated cyber-attacks.

\newpage

\bibliographystyle{acl_natbib}
\bibliography{main}

\input{sec/supp}

\end{document}

%% file: preamble.tex
\usepackage{booktabs} 
\usepackage{amsmath}  
\usepackage{geometry} 

\usepackage{xcolor}     
\usepackage{amssymb}    
\usepackage{makecell}   

\usepackage{cleveref}

\usepackage{multirow}

\usepackage{colortbl}

\usepackage[most]{tcolorbox} 
\usepackage{listings}        

\newcommand{\diff}[2]{\rlap{$_{\textcolor{#1}{#2}}$}}

\definecolor{boxheader}{RGB}{220, 53, 69} 
\definecolor{boxbg}{RGB}{255, 248, 248}   
\definecolor{boxframe}{RGB}{180, 40, 55}  
\definecolor{subhead}{RGB}{0, 0, 139}     

\newtcolorbox{promptbox}[2][]{
    enhanced,
    breakable,
    colback=boxbg,
    colframe=boxframe,
    coltitle=white,
    colbacktitle=boxheader,
    fonttitle=\bfseries\large,
    title={#2},
    arc=2mm,
    boxrule=0.5mm,
    titlerule=0mm,
    top=4mm, bottom=4mm, left=4mm, right=4mm,
    drop fuzzy shadow,
    #1
}

\newcommand{\promptsection}[1]{\vspace{0.5em}\noindent\textcolor{subhead}{\textbf{\large #1}}\par}

\newcommand{\cmark}{\textcolor{green!60!black}{\ensuremath{\checkmark}}} 
\newcommand{\xmark}{\textcolor{red}{\ensuremath{\times}}}

%% file: sec/abstract.tex
\vspace{20pt}

\begin{abstract}


Large Language Models (LLMs) based autonomous agents demonstrate multifaceted capabilities to contribute substantially to economic production. However, existing benchmarks remain focused on single agentic capability, failing to capture long-horizon real-world scenarios. 
Moreover, the reliance on human-in-the-loop feedback for realistic tasks creates a scalability bottleneck, hindering automated rollout collection and evaluation. To bridge this gap, we introduce \textsc{AgencyBench}, a comprehensive benchmark derived from daily AI usage, evaluating 6 core agentic capabilities across 32 real-world scenarios, comprising 138 tasks with specific queries, deliverables, and rubrics. These scenarios require an average of \textbf{90 tool calls, 1 million tokens, and hours of execution time} to resolve. To enable automated evaluation, we employ a user simulation agent to provide iterative feedback, and a Docker sandbox to conduct visual and functional rubric-based assessment. Experiments reveal that closed-source models significantly outperform open-source models (48.4\% vs 32.1\%). Further analysis reveals significant disparities across models in resource efficiency, feedback-driven self-correction, and specific tool-use preferences. Finally, we investigate the impact of agentic scaffolds, observing that proprietary models demonstrate superior performance within their native ecosystems (e.g., Claude-4.5-Opus via Claude-Agent-SDK), while open-source models exhibit distinct performance peaks, suggesting potential optimization for specific execution frameworks. \textsc{AgencyBench} serves as a critical testbed for next-generation agents, highlighting the necessity of co-optimizing model architecture with agentic frameworks. We believe this work sheds light on the future direction of autonomous agents, and to facilitate community adoption, we release the full benchmark and evaluation toolkit at \href{https://github.com/GAIR-NLP/AgencyBench}{https://github.com/GAIR-NLP/AgencyBench}.

\end{abstract}






%% file: fig/fig_tex/teaser.tex
\vspace{20pt}

\begin{center}
    \noindent
    \begin{minipage}[c]{0.39\linewidth} 
        \centering
        \includegraphics[width=\linewidth]{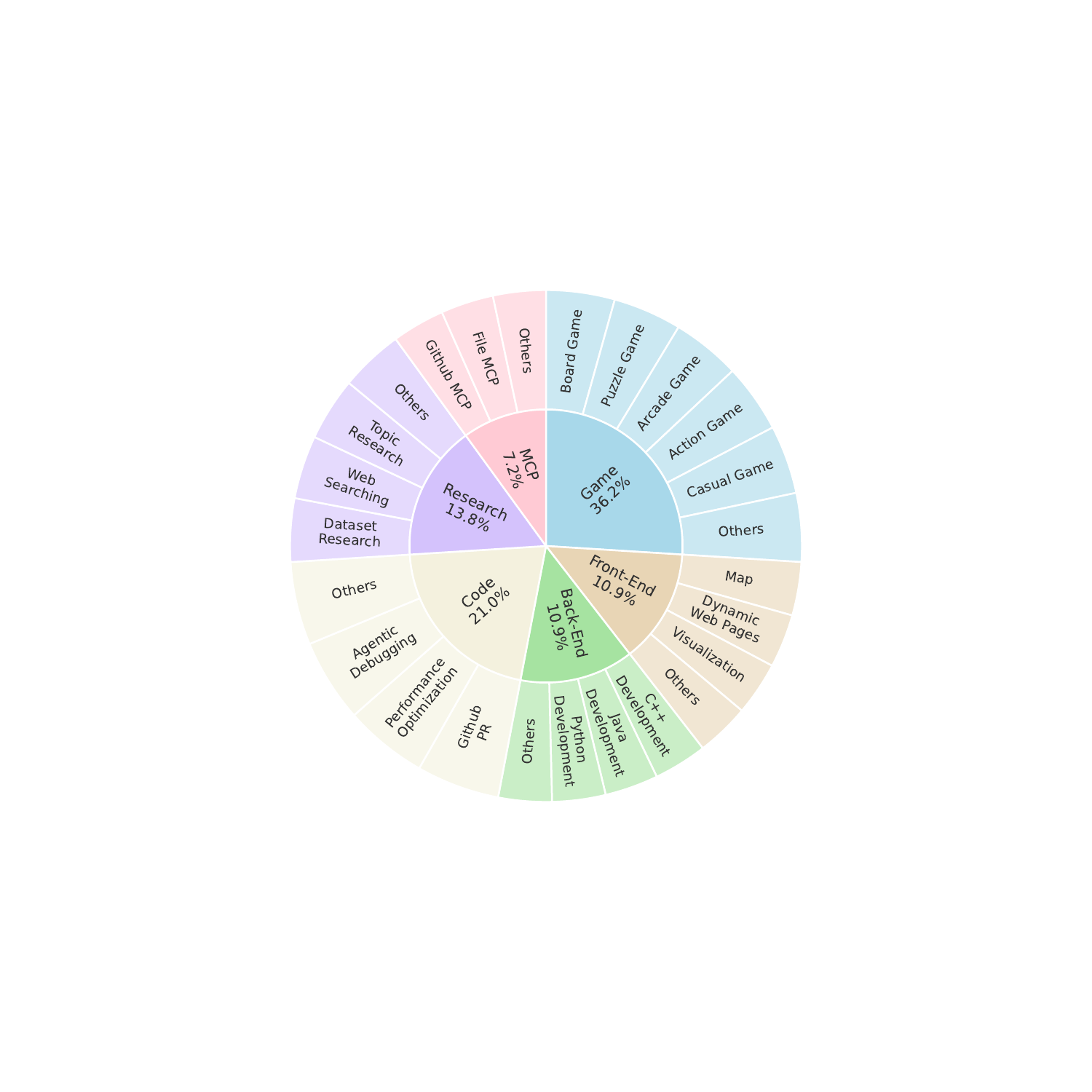}
    \end{minipage}%
    \hfill 
    \begin{minipage}[c]{0.6\linewidth} 
        \centering
        \resizebox{0.95\linewidth}{!}{%
        \scriptsize 
            \setlength{\tabcolsep}{2pt} 
            \begin{tabular}{l|ccccc}
            \toprule
            \textbf{Benchmark} &
            \makecell[c]{\textbf{Avg.} \\ \textbf{Tok.}} &
            \makecell[c]{\textbf{Avg.} \\ \textbf{Turns}}&
            \makecell[c]{\textbf{Diverse} \\ \textbf{Agentic}} &
            \makecell[c]{\textbf{User} \\ \textbf{Sim.}} &
            \makecell[c]{\textbf{Docker} \\ \textbf{Sandbox}} \\
            \midrule
            Browsecomp & -- & -- & \xmark & \xmark & \xmark \\
            Terminal-bench & -- & -- & \xmark & \xmark & \xmark \\
            SWE-verified & -- & 15 & \xmark & \xmark & \cmark \\
            MCPUniverse & -- & 7.5 & \xmark & \xmark & \xmark \\
            GAIA2 & 10K & 22.5 & \cmark & \xmark & \xmark \\
            Toolathlon & 15K & 26.8 & \cmark & \xmark & \cmark \\
            UltraHorizon & 200K & 60 & \cmark & \xmark & \xmark \\
            \midrule
            \textbf{\textsc{AgencyBench}} & \textbf{1M} & \textbf{90} & \cmark & \cmark & \cmark \\ 
            \bottomrule
            \end{tabular}%
        }
    \end{minipage}

    \vspace{5pt} 

\captionof{figure}{\textbf{Overview of \textsc{AgencyBench}.} \textbf{Left:} Distribution of the 32 scenarios and 138 tasks across 6 distinct agentic capabilities. \textbf{Right:} Comparison with existing benchmarks. \textsc{AgencyBench} focuses on diverse, long-horizon real-world tasks, requiring an average of 1M tokens and 90 multi-turn tool uses. It integrates a user simulation agent for iterative feedback and a Docker-based sandbox for automated rubric-based assessment.}
    \label{fig:teaser}
\end{center}

%% file: sec/introduction.tex
\section{Introduction}



With the rapid advancement of Large Language Models (LLMs) \cite{gpt5.2,gemini3,claude4.5,grok4.1}, integrating these models with advanced scaffolds to form autonomous agents has become a paradigm shift. As these agents increasingly permeate diverse domains, ranging from economic production, scientific research, and software development to everyday use, establishing rigorous benchmark to measure their practical economic value and performance is becoming unprecedentedly urgent \cite{pan2025measuring,openrouter,kwa2025measuring}. However, current agent benchmarks face significant limitations: (1) Existing benchmarks often exhibit a scarcity of long-horizon tasks~\cite{li2025tool,andrews2025scaling} or focus narrowly on single agentic capability, such as tool use~\cite{terminalbench,wu2025mcpmark}, software engineering~\cite{jimenez2023swe} or research~\cite{wei2025browsecomp,xu2025researcherbench}, failing to capture the long-horizon nature and diversity of real-world tasks. (2) Furthermore, completing realistic tasks often necessitates continuous human feedback to guide agents through multi-turn interactions. This reliance on human-in-the-loop processes creates a bottleneck, restricting the automation of rollout collection and evaluation.

To bridge this gap, we introduce \textsc{AgencyBench}, a comprehensive benchmark designed to evaluate agent capabilities through highly long-horizon, diverse, and authentic real-world tasks. By requesting AI researchers, active AI practitioners, and software engineer developers, we systematically construct 32 real-world scenarios evaluating 6 core agentic capabilities, comprising a total of 138 specific tasks. Each task is defined by queries (descriptions of task requirements), deliverables (descriptions of the expected outputs), and rubrics (evaluation criteria for assessment). These scenarios are notably demanding: on average, resolving a single scenario approximately \textbf{90 tool calls, consumes 1 million tokens, and requires hours of execution time}, rigorously evaluating agents' ability to maintain context and execute logic over extended periods to satisfy multi-turn, long-horizon real-world needs. The comparison between \textsc{AgencyBench} and various other benchmarks is shown in \Cref{fig:teaser}.


To facilitate scalable and automated rollout collection, we develop an agent scaffold equipped with a comprehensive tool suite operating within an isolated \textit{workspace}. In this environment, the agent engages in multi-turn interactions to generate raw deliverables, assisted by a user simulation agent that mimics real-world scenarios and provides iterative feedback to bypass human-in-the-loop limitations. Subsequently, these deliverables are synchronized to a Docker-based remote sandbox, which emulates human-computer operations (e.g., UI rendering, mouse/keyboard inputs) to produce visual evaluation artifacts. The process concludes in a separate \textit{eval-space}, where these artifacts and raw deliverables undergo automated rubric-based assessment using executable scripts.

Extensive experiments on \textsc{AgencyBench} indicate that closed-source models achieve an average score of 48.4\%, while open-source models average 32.1\%. Closed-source models range from 56.5\% (GPT-5.2) to 44.3\% (Grok-4.1-Fast), whereas open-source models span from 38.6\% (GLM-4.6) to 27.0\% (Qwen-3-235B-A22B-Thinking). The overall performance reveals that current frontier models still struggle to master the long-horizon, real-world tasks. 
Further analysis reveals distinct behavioral differences among the models: GPT-5.2 demonstrates stronger capabilities in feedback-driven self-correction, Grok-4.1-fast exhibits higher token utilization efficiency, Claude-4.5-Opus shows a preference for shell-based tools, and Gemini-3-Pro favors file-related and memory management tools. Additionally, comparative studies on agentic scaffolds highlight a `home-field advantage', where models achieve peak performance when paired with their native or specifically optimized frameworks.

In summary, our contributions are as follows: (1) We introduce \textsc{AgencyBench}, a challenging benchmark with 138 authentic tasks across 32 scenarios to evaluate long-horizon agentic capabilities. (2) We develop a unified evaluation framework leveraging user simulation agent and Docker-based remote sandboxes to achieve fully automated evaluation. (3) We provide a comprehensive evaluation of frontier models, quantifying the gap between proprietary and open-source models and uncovering distinct behavioral patterns.

%% file: sec/related.tex
\input{fig/fig_tex/bench_all}

\section{Related Work}


\paragraph{Large Language Model Agents} The continuous and rapid evolution of large language models (LLMs) \cite{team2025kimi,yang2025qwen3,liu2025deepseek,glm4.6,grok4.1,claude4.5,gemini3,gpt5.2} has fundamentally reshaped the landscape of artificial intelligence, propelling the field from simple conversational tasks to complex reasoning \cite{xiao2025limopro,xiao2025scale,xiao2025towards} and multi-turn tool-use tasks \cite{li2025datasetresearch,xiao2025limi,wu2025innovatorbench}. To enable these models to tackle intricate real-world challenges, numerous agentic scaffolds have been developed \cite{yao2022react,yang2024swe,wang2024openhands,cursor-cli,codex,claudecode}, effectively unlocking the potential for long-horizon task completion through iterative feedback and structured planning. Agents are now expected to autonomously navigate diverse scenarios, ranging from full-stack front-end development to complex game engine manipulations. 

\paragraph{Agent Benchmarks} To rigorously evaluate these growing capabilities, various benchmarks have emerged, focusing on specific vertical domains like tool use \cite{li2025tool,barres2025tau,terminalbench,wu2025mcpmark}, software development \cite{miserendino2502swe,designarena,jimenez2023swe}, or open-ended research \cite{wei2025browsecomp,andrews2025scaling,patwardhan2025gdpval}, with some recent efforts assessing potential economic utility \cite{miserendino2502swe,patwardhan2025gdpval,xiao2025towards}. However, current benchmarks often lack the necessary complexity to differentiate frontier models, as the limited number of tool calls and shallow interaction depths remain insufficient to test the upper limits of modern agents \cite{wu2025mcpmark,andrews2025scaling,li2025tool,luo2025ultrahorizon}. \textsc{AgencyBench} addresses this critical gap by curating 138 authentic, high-fidelity tasks across 6 agentic capabilities, where the average rollout exceeds 1M tokens and necessitates over 90 precise tool calls, significantly raising the bar for task difficulty and context retention.

%% file: fig/fig_tex/bench_all.tex
\begin{figure*}[t]
    \centering
\includegraphics[width=1.0\linewidth]{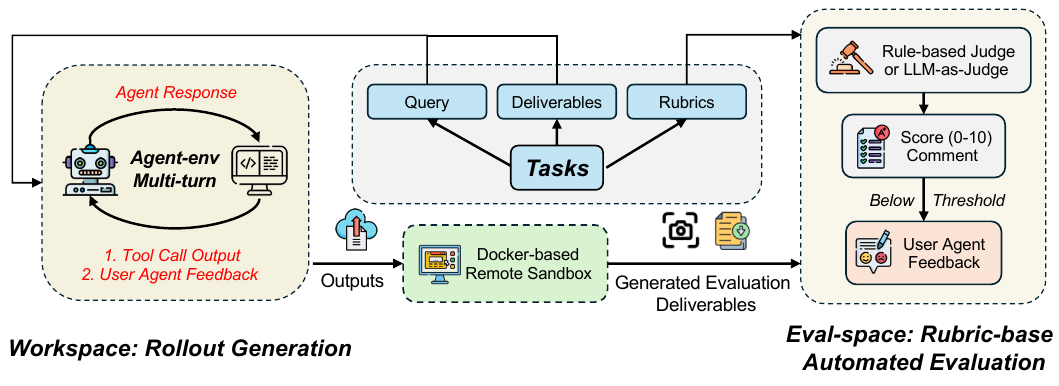}

\caption{\textbf{\textsc{AgencyBench} Rollout Generation and Evaluation Pipeline.} 
    Rollout generation takes place within \textit{workspace}, where the agent receives task queries and deliverables, completing tasks through multi-turn interactions with the environment (e.g., tool execution results and feedback from the user simulation agent). Upon task completion, deliverables are synced to a Docker sandbox for operation execution (e.g., UI actions), and resulting artifacts are transferred to \textit{eval-space} for scoring ($0-10$) via rule-based or LLM-based judges based on task rubrics.}
    \label{fig:pipeline}
\label{fig:bench_all}
\end{figure*}





%% file: sec/method.tex
\section{\textsc{AgencyBench}}
In this section, we detail the hierarchical design of \textsc{AgencyBench} (\Cref{subsec:design}), the data collection process and the formal definition of interaction rollouts (\Cref{subsec:rollout}). Finally, we describe our automated evaluation framework (\Cref{subsec:evaluation}), which leverages a Docker-based remote sandbox and rubric-based rigorous, scalable, and reproducible assessment. Rollout generation and evaluation pipeline is illustrated in \Cref{fig:bench_all}, with a specific evaluation example detailed in \Cref{fig:bench_case}.

\subsection{\textsc{AgencyBench} Design}
\label{subsec:design}

\subsubsection{Design Pattern}

\paragraph{Capabilities, Scenarios, and Tasks}
To capture the multifaceted and long-horizon nature of real-world tasks, \textsc{AgencyBench} is structured hierarchically. It evaluates 6 core agentic capabilities: game development, front-end development, back-end development, code generation, research, and MCP tool use. These capabilities are distributed across 32 authentic real-world scenarios, such as developing a Gomoku game from scratch (developing board game capability), conducting project-level code debugging (agentic debugging capability), performing in-depth corporate research (research capability), etc. Serving as a complete testing unit, each scenario is structured as a logical hierarchy of 1 to 5 tasks, arranged in ascending order of difficulty and presented sequentially, where the completion results of preceding tasks influence subsequent ones. Through this design, scenarios are expanded into a comprehensive collection totaling 138 distinct tasks. This design is intended to simulate real-world tasks that progress from basic to complex tasks, thereby requiring agent to maintain context and execute logic over extended periods to satisfy multi-turn, long-horizon nature of real-world tasks.

\paragraph{Workspace, Eval-space and Scaffold}
\label{para:scaffold}
To address the limitations of existing benchmarks that rely on unscalable human experts' feedback for complex environments, we design a robust execution infrastructure. Each task operates within an isolated \textit{workspace} to ensure reproducibility and prevent state interference. This workspace is equipped with an agent scaffold (a suite of tools including file manipulation, command-line execution, web search, context management, ...) allowing the agent to generate rollouts in a realistic setting. 
For each task, the agent engages in multi-turn interactions with the environment using the scaffold within the \textit{workspace} to generate deliverables.
To facilitate assessment, deliverables are synced to a Docker-based remote sandbox, which emulates human computer operations (e.g., mouse clicks, UI rendering) to produce visualizable artifacts. These artifacts are subsequently transferred to a local \textit{eval-space}, where the evaluation process is completed automatically using executable scripts without human intervention.

\subsubsection{Data Collection}
We employ 20 human experts, including AI researchers, active AI practitioners, software engineer developers to collect real-world tasks. Based on this data, these human experts systematically construct the 32 scenarios and 138 tasks. For each task, experts manually construct and verify three key components: (1) the \textbf{Query}, which describes the specific requirements; (2) the \textbf{Deliverables}, which define the expected file outputs or terminal states; and (3) the \textbf{Rubrics}, which establish the objective criteria for assessment. Furthermore, experts develop executable evaluation scripts for each task. A separate panel of four experts conducts a comprehensive review of the entire dataset to verify descriptive accuracy, difficulty calibration, and environment configurations. To ensure the highest quality standards, a strict unanimous consensus policy is enforced: a task is only finalized if all experts reach a full agreement. If any discrepancy arises, the task is flagged and remanded for revision, requiring subsequent re-evaluation until it meets the 100\% approval threshold.

\input{fig/fig_tex/bench_case}

\subsection{Rollout Collection}
\label{subsec:rollout}

\paragraph{Rollout Definition}
We formalize the interaction process as a sequence of states and actions. Assuming a scenario consists of five sequential tasks, we denote their individual rollouts as $\tau_1$, $\tau_2$, ... $\tau_5$. These are defined as follows:

$$
\begin{aligned}
\tau_1 &= (q_1, a, t, \dots, a, u_{11}, \dots, a, u_{12}, ..., a, t, ...) \\
\tau_2 &= (q_2, a, t, \dots, a, u_{21}, \dots, a, u_{22}, ..., a, t, ...) \\
... \\
\tau_5 &= (q_5, a, t, \dots, a, u_{51}, \dots, a, u_{52}, ..., a, t, ...) \\
\end{aligned}
$$

Here, for the $i$-th task, $q_i$ represents the initial query provided by \textsc{AgencyBench}. The pair $(a, t)$ denotes the agent's reasoning and specific tool calls (e.g., file writing, shell commands). 
A user simulation agent provides feedback $u_{ij}$ (which prompts further agent actions) whenever the deliverables fail to meet the score threshold prescribed by the rubric. The complete rollout $\tau$ for the scenario is defined as the ordered concatenation of these trajectories:

$$
\begin{aligned}
\tau = (\tau_1, \tau_2, \tau_3, \tau_4, \tau_5)
\end{aligned}
$$

This formulation captures the iterative and long-horizon nature of real-world tasks.

\paragraph{User Simulation Agent}
To simulate realistic human-agent collaboration without manual intervention, we implement a user simulation agent, responsible for providing feedback to the task executing agent, enabling targeted improvements based on task execution. Feedback is determined based on the fulfillment of the current task's rubrics. For instance, if an agent completes only 6 out of 10 rubrics, the user simulation agent will return the 4 failed rubrics along with their specific reasons for failure. We utilize Claude-4-Sonnet with a temperature setting of 0.0 for this role (detailed prompts in \Cref{sec:prompts}). 

To ensure the validity of the user simulation agent, we conducted a human verification study on 50 randomly sampled interaction rollouts. Four human experts independently drafted justifications for unmet rubrics, which another four experts then independently scored against the agent's feedback. Scores were assigned on an integer scale from 0 to 5, representing `Highly Inconsistent', `Inconsistent', `Uncertain', `Consistent', and `Highly Consistent', respectively. The final average score reached 4.69, demonstrating a high degree of alignment and confirming that the agent serves as a reliable surrogate for human experts.

\input{tab/main_exp_1}

\subsection{Evaluation}
\label{subsec:evaluation}
Our evaluation framework is entirely rubric-based, ensuring standardized assessment across diverse domains. We employ executable evaluation scripts that map the agent's deliverables to a score ranging from 0 to 10. Depending on the nature of the task, we utilize either rule-based methods or LLM-as-judge mechanisms.

\paragraph{Rule-based Evaluation}
For tasks with objectively verifiable ground truth, such as correct tool execution, mathematical optimization, or specific file generation—we employ rule-based evaluation. In these cases, the rubrics are directly translated into assertion logic within the evaluation scripts. The final score is calculated by mapping the pass rate of these assertions or the optimization metric to the 0-10 scale.

\paragraph{LLM-as-Judge}
For tasks involving subjective qualities or complex visual outputs (e.g., game aesthetics, front-end layout), we employ an LLM-as-judge. 
Specifically for game and frontend scenarios, we implement a multimodal judging system: 
\textbf{(1) Text-based Judge:} Evaluates code quality and logic based on the text deliverables and rubrics. We utilize Claude-4-Sonnet with a temperature of 0.0.
\textbf{(2) Vision-based Judge:} Evaluates dynamic behavior and visual correctness based on screenshots and recordings captured from remote sandbox interactions. We utilize Gemini-2.5-pro with a temperature of 0.0. Detailed prompts are provided in \Cref{sec:prompts}. 

For tasks requiring visual deliverables, the final score is calculated as the average of the ratings from the text and vision-based judges; otherwise, the score is determined solely by the text-based judge. To validate the reliability of our LLM judges, we compare their ratings against human annotations on a held-out set of 50 tasks. Four human experts independently scored these 50 tasks based on the rubrics. The results showed a Kappa score of 0.93 between the human and LLM judge scores, demonstrating the reliability of the evaluation.

To comprehensively evaluate agent capabilities and resource use in \textsc{AgencyBench}, we adopt the following metrics for assessment.

\paragraph{Metric: Average Score ($S_{Avg}$)}
Calculated as the percentage derived from rubric-based evaluations (e.g., satisfying 6 out of 10 criteria yields 60\%), where higher scores indicate superior capability.

\paragraph{Metric: Average Attempts ($Att$)}
Measures the average iterations per scenario. A task is passed if the task score is at least $60\%$. If a task fails, the user simulation agent provides feedback up to a maximum of $K$ rounds until the threshold is met. $Att$ denotes the average rounds used (ranging from 1 to $K$), where lower values imply stronger autonomy.
Let $M$ denote the total number of tasks in the current scenario, and $M_{Att}$ represent the total count of attempts used across all tasks. The Average Attempts ($Att$) is defined as:

\begin{equation}
    Att = \frac{M_{Att}}{M}
\end{equation}

\paragraph{Metric: Pass Rate ($Pass@k$)}
Given $N$ tasks in a scenario, let $N_{pass}$ be the number of tasks achieving the 60\% score threshold within $k$ feedback rounds. The metric is defined as:

\begin{equation}
Pass@k = \frac{N_{pass}}{N} 
\end{equation}

We set the feedback limit to 1 and 2, reporting $Pass@1$ (up to 1 feedback round) and $Pass@2$ (up to 2 feedback rounds), respectively.

\paragraph{Metric: Efficiency}
We denote $Tok$ as average tokens used per scenario and compute \textit{Attempt Efficiency} ($E_{att}$) and \textit{Token Efficiency} ($E_{tok}$) to normalize performance against resource costs where higher values  indicate better resource optimization:

\begin{equation}
    E_{att} = \frac{S_{avg}}{Att}, \quad E_{tok} = \frac{S_{avg}}{Tok}
\end{equation}

%% file: fig/fig_tex/bench_case.tex
\begin{figure*}[t]
    \centering
\includegraphics[width=1.0\linewidth]{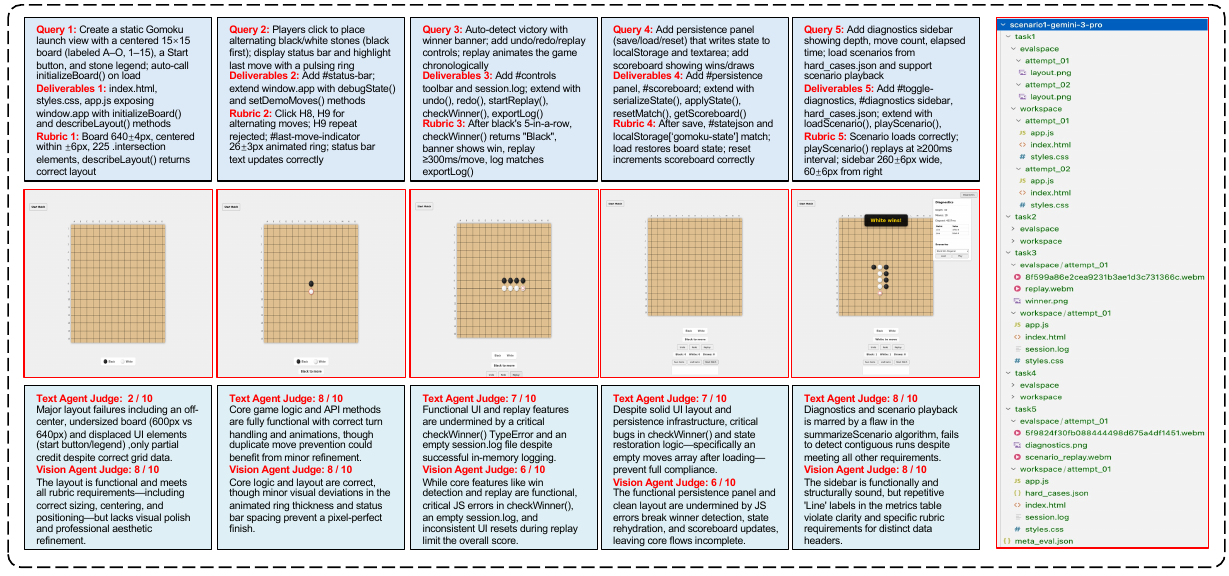}

\caption{\textbf{An Illustrative Evaluation Scenario in \textsc{AgencyBench}: Developing a Gomoku Game.} 
    The scenario consists of five sequential tasks with increasing complexity, requiring the incremental addition of new features. The primary deliverables include HTML, CSS, and JS source code. Evaluation scripts execute these files within a remote Docker sandbox, performing interactive operations such as clicking, screen recording, and capturing screenshots (visualized as video frames in the figure). The resulting evaluation artifacts are retrieved to \textit{eval-space}, where text and vision agents assess the code and visual deliverables, respectively, providing scores and qualitative feedback based on rubrics. 
    \textbf{Right:} The file organization architecture during runtime, showing the isolated workspace and eval-space for each task to ensure environmental consistency and prevent cross-task interference.}
    
\label{fig:bench_case}
\end{figure*}

%% file: tab/main_exp_1.tex
\begin{table*}[t]
\centering
\small 
\setlength{\tabcolsep}{3.5pt} 

\begin{tabular}{lccccc c<{\hspace{1.8em}}@{} | c<{\hspace{1.8em}}@{} c}
\toprule
Model & Game $\uparrow$ & Frontend $\uparrow$ & Backend $\uparrow$ &  Code $\uparrow$ & Research $\uparrow$ & MCP $\uparrow$ & $\boldsymbol{S_{Avg} \uparrow}$ & $\boldsymbol{Att \downarrow}$ \\
\midrule
\multicolumn{8}{c}{\textit{Proprietary Models}} \\

\midrule

\rowcolor{cyan!20} 
GPT-5.2            
& 52.2 & 74.7 & 61.0 & 50.7 & 64.4 & 52.1 & 56.5 & 1.46 \\

Claude-4.5-Opus 
& 52.1\diff{blue}{-0.1} & 49.3\diff{blue}{-25.4} &  49.6\diff{blue}{-11.4} & 24.0\diff{blue}{-26.7} & 68.8\diff{red}{+4.4} & 64.4\diff{red}{+12.3}  &  47.7\diff{blue}{-8.8} & 1.54 \\

Gemini-3-Pro        
& 60.7\diff{red}{+8.5} & 81.0\diff{red}{+6.3} &  31.3\diff{blue}{-29.7} & 23.5\diff{blue}{-27.2} & 40.0\diff{blue}{-24.4} & 61.8\diff{red}{+9.7}  & 46.9\diff{blue}{-9.6} & 1.46 \\

Claude-4.5-Sonnet   
& 56.5\diff{red}{+4.3} & 51.6\diff{blue}{-23.1} &  35.3\diff{blue}{-25.7} & 17.3\diff{blue}{-33.4} & 71.4\diff{red}{+7.0} & 62.3\diff{red}{+10.2}  & 46.4\diff{blue}{-10.1} & 1.49 \\

Grok-4.1-fast  
& 38.8\diff{blue}{-13.4} & 65.7\diff{blue}{-9.0} &  37.3\diff{blue}{-23.7} & 26.4\diff{blue}{-24.3} & 63.8\diff{blue}{-0.6} & 68.6\diff{red}{+16.5}  & 44.3\diff{blue}{-12.2} & 1.55 \\

\midrule

\multicolumn{8}{c}{\textit{Open-Source Models}} \\

\midrule

\rowcolor{cyan!20} 
 GLM-4.6   
 & 59.2 & 64.3 & 20.0 & 11.9 & 32.0 & 49.8  & 38.6 & 1.54 \\

 Kimi-K2-Thinking       
& 40.6\diff{blue}{-18.6}  &  54.7\diff{blue}{-9.6} & 24.3\diff{red}{+4.3} & 11.8\diff{blue}{-0.1}  & 33.6\diff{red}{+1.6}  & 31.4\diff{blue}{-18.4}  & 34.2\diff{blue}{-4.4} & 1.79 \\
 
Deepseek-V3.2          
& 36.5\diff{blue}{-22.7} &  49.3\diff{blue}{-15.0} & 20.7\diff{red}{+0.7} & 11.6\diff{blue}{-0.3} & 22.0\diff{blue}{-10.0} & 48.5\diff{blue}{-1.3}  & 28.6\diff{blue}{-10.0} & 1.63 \\

Qwen-3-235B-A22B-Thinking        
& 40.4\diff{blue}{-18.8}  &  57.7\diff{blue}{-6.6} & 3.3\diff{blue}{-16.7} & 4.6\diff{blue}{-7.3} & 34.4\diff{red}{+2.4} & 20.9\diff{blue}{-28.9}  & 27.0\diff{blue}{-11.6} & 1.79 \\

\bottomrule
\end{tabular}
\caption{\textbf{Main Experimental Results.} The table compares proprietary and open-source models. The rows for GPT-5.2 and GLM-4.6 are highlighted in blue as baselines. Colored subscripts indicate the performance gap compared to the baseline (\textcolor{red}{red} for improvement, \textcolor{blue}{blue} for degradation).}

\label{tab:main_exp_1}
\end{table*}

%% file: sec/exp.tex
\input{fig/fig_tex/effieiency}

\section{Experiments}

In this section, we conduct a systematic evaluation of various LLMs on \textsc{AgencyBench}. We first outline experimental setup and metrics (\Cref{subsec:setup}), followed by analysis of overall performance across agentic capabilities (\Cref{subsec:main_result}). We then investigate the crucial ability of self-correction through feedback (\Cref{subsec:self}), analyze resource use (\Cref{subsec:resourse}) and the economic efficiency of models (\Cref{subsec:efficiency}), and decode the distinct behavioral patterns in tool usage (\Cref{subsec:tool}). Finally, we examine the influence of different agentic scaffolds on model performance (\Cref{subsec:scaffold_exp}).

\subsection{Experimental Setup}
\label{subsec:setup}

\paragraph{Models and Scaffold}
We evaluate a comprehensive suite of models, comprising closed-source LLMs: GPT-5.2~\cite{gpt5.2}, Claude-4.5-Opus, Claude-4.5-Sonnet~\cite{claude4.5}, Grok-4.1-Fast~\cite{grok4.1} and open-source LLMs: GLM-4.6~\cite{glm4.6}, DeepSeek-V3.2~\cite{liu2025deepseek}, Qwen-3-235B-A22B-Thinking~\cite{yang2025qwen3}, Kimi-K2-Thinking~\cite{team2025kimi}. All models are accessed via OpenRouter\footnote{https://openrouter.ai/} API with the temperature set to 0.7 to balance creativity and determinism. We utilize the agentic scaffold described in \Cref{para:scaffold}, which is equipped with a robust set of tools (detailed in \Cref{tab:each_tool}).

\input{tab/attempt_1}

\subsection{Main Results}
\label{subsec:main_result}

\paragraph{Overall Performance and Efficiency}
Table \ref{tab:main_exp_1} highlights a capability gap between proprietary and open-source models. GPT-5.2 achieves the highest overall average score (56.5\%) among proprietary models, whereas GLM-4.6 leads the open-source category with 38.6\%. Conversely, Qwen-3-235B-A22B-Thinking records the lowest performance (27.0\%), underscoring that substantial room for improvement.
Regarding user simulation agent feedback attempts (Att), GPT-5.2 and Gemini-3-Pro demonstrate superior capability, requiring the fewest average attempts (1.46) to complete tasks. In the open-source sector, GLM-4.6 performs comparably to proprietary models (1.54), while Qwen-3-235B-A22B-Thinking and Kimi-K2-Thinking struggle with higher attempt counts (1.79), indicating weaker error-recovery abilities.

\input{tab/resource_1}

\paragraph{Agentic Capabilities}
Performance varies significantly across agentic capabilities, revealing distinct specializations. Gemini-3-Pro dominates game (60.7\%) and front-end (81.0\%). GPT-5.2 excels in back-end and code, while Claude-4.5-Sonnet achieving the highest in research (71.4\%). Among open-source models, GLM-4.6 exhibits balanced performance, while Qwen-3-235B-A22B-Thinking demonstrates relative strength in research despite lower overall average.

\subsection{Feedback-driven Self-correction Analysis}
\label{subsec:self}
\Cref{tab:attempt_1} quantifies feedback-driven self-correction ability by \textit{Pass@1} and \textit{Pass@2}.
Top-tier proprietary models demonstrate sophisticated error handling. GPT-5.2 achieves an 88.9\% relative increase, and the Claude series similarly exceeds 80\% improvement. In contrast, while Gemini-3-Pro matches the initial Pass@1 of GPT-5.2, its ability to leverage feedback is markedly lower (33.3\% Rise), suggesting that while its initial intuition is strong, its self-correction mechanisms are less responsive.
In the open-source domain, Kimi-K2-Thinking and Qwen-3-235B-A22B-Thinking achieve remarkable improvements after feedback (300\% and nearly 200\%, respectively). Conversely, DeepSeek-V3.2 achieves (0.0\% Rise), persistently adhering to erroneous paths despite external critique.

\subsection{Resource Consumption Analysis}
\label{subsec:resourse}
\Cref{tab:resource_1} details the trade-off between performance ceilings and costs: Tok represents the token consumption, measured in millions; T represents the average scenario execution time, measured in hours; Turns represents the number of tool-calling rounds.
GPT-5.2 acts as a `brute-force' reasoner, consuming 3.4 million tokens and 89 turns on average to secure top scores. In contrast, Grok-4.1-Fast represents the pinnacle of speed and frugality (1.2M tokens, 0.3h). GLM-4.6 exhibits a unique profile: despite a high turns count (105), its resource usage remains moderate. Notably, Kimi-K2-Thinking and Qwen-3-235B-A22B-Thinking incur the highest time costs (1.2h), likely due to the latency of generating internal reasoning traces.

\input{fig/fig_tex/each_tool_1}

\input{tab/scaffold_main}

\subsection{Attempt and Token Efficiency Analysis}
\label{subsec:efficiency}
To decouple raw performance from expenditure, we analyze efficiency in \Cref{fig:efficiency}.
\textit{Attempt Efficiency} is dominated by GPT-5.2 (38.7\%), indicating that its high resource usage is justified by a high success rate per attempt. \textit{Token Efficiency} favors Grok-4.1-Fast (37.2\%), making it the most economically viable choice for resource-constrained environments. Claude-4.5-Sonnet ranks lowest (11.4\%), indicating excessive token generation (4.1M) does not yield proportional performance gains.

\subsection{Behavioral Patterns in Tool Invocation}
\label{subsec:tool}
\Cref{fig:each_tool_1} reveals model architectures imprint distinct `personalities' on problem-solving strategies:
Claude-4.5-Opus and GPT-5.2 prefer system-level manipulation via shell execution (45.5\% and 43.5\%); Gemini-3-Pro distinctively utilizes explicit memory tools (6.9\%), suggesting a strategy rooted in managing long-horizon context banks; Qwen-3-235B-A22B-Thinking exhibits an extreme reliance on file operations (77.6\%), prioritizing direct content verification; Grok-4.1-Fast and GLM-4.6 exhibit a high reliance on web search (9.5\% and 8.6\%), appearing to offload knowledge retrieval to external sources rather than relying on internal parametric memory.

\subsection{Impact of Agentic Scaffolds}\label{subsec:scaffold_exp}To investigate the influence of agentic scaffold on model performance, we conducted an ablation study on a subset of 10 representative scenarios using three distinct frameworks: our native scaffold (used in the main experiments), the Claude-Agent-SDK~\cite{claude-code-sdk}, and the OpenAI-Agents-SDK~\cite{openai-agent-sdk}. We report the Average Score ($S_{Avg}$) in \Cref{tab:scaffold_main}. 

\paragraph{Ecosystem Synergy in Proprietary Models} Experimental results reveal a significant `Ecosystem Synergy' effect, where proprietary models demonstrate peak performance within their native frameworks. Most notably, Claude-4.5-Opus achieves a substantial performance boost of 20.5\% when operating within the Claude-Code SDK compared to our generalist scaffold. This suggests a deep optimization between the model's training objective and its proprietary tool definitions/prompt structures. Similarly, GPT-5.2 shows a preference for the OpenAI-Agents-SDK (+1.3\%), outperforming its results on third-party alternatives.

\paragraph{Scaffold Sensitivity in Open-Source Models}
Among open-source models, the impact of scaffold choice is heterogeneous. GLM-4.6 and Minimax-M2 exhibit strong compatibility with the Claude-Agent-SDK scaffold, improving by 10.6\% and 8.6\% respectively. This improvement may stem from the SDK's structured prompt engineering, which likely aligns well with the instruction-following capabilities of these models. \textbf{Furthermore, it suggests the possibility that these models may have been specifically optimized during training to adapt to the interaction patterns characteristic of the Claude-Agent-SDK ecosystem.} Conversely, Kimi-K2-Thinking performs best on our custom scaffold, experiencing significant degradation when migrated to external SDKs (dropping by 12.8\% on the OpenAI-Agents-SDK). These findings underscore that agentic performance is not solely a model-intrinsic property but a result of the coupling between the model and its agentic scaffold.

%% file: fig/fig_tex/effieiency.tex
\begin{figure*}[t]
    \centering
\includegraphics[width=1.0\linewidth]{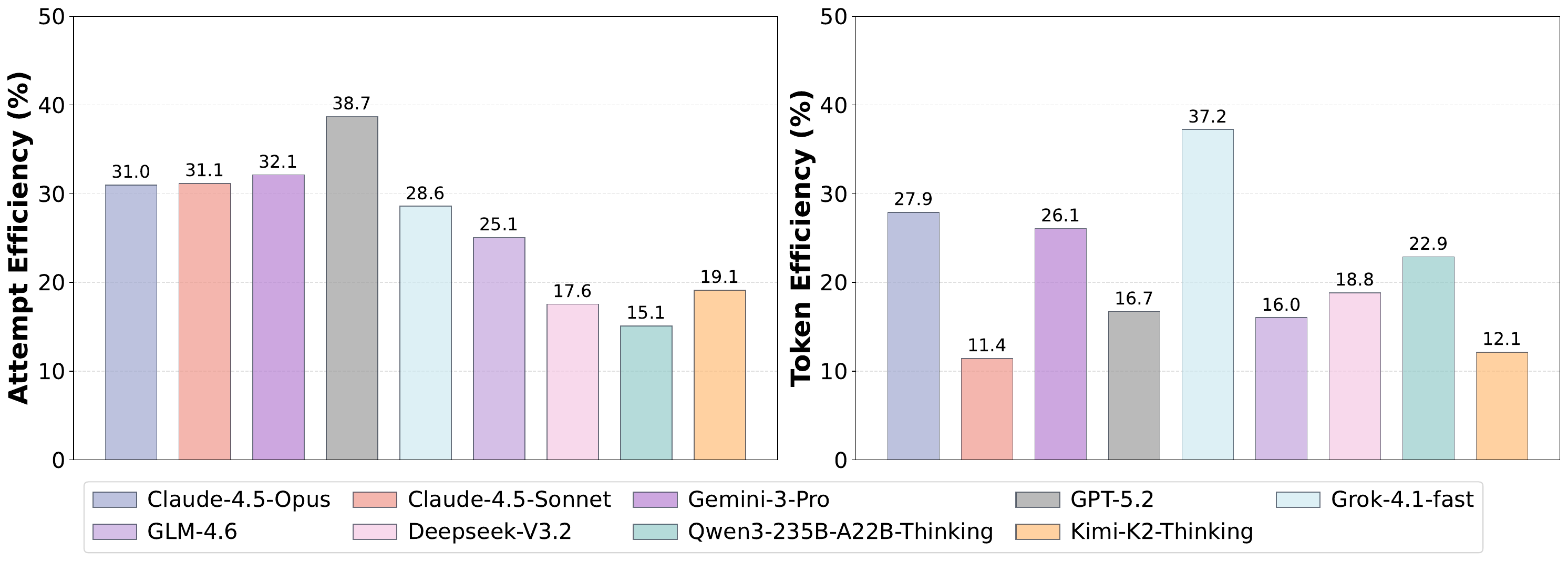}

\caption{\textbf{Efficiency Comparison Across Models.} Efficiency is calculated by dividing the average score by the number of attempts and average token consumption, respectively. GPT-5.2 achieves the highest attempt efficiency, while Qwen-3-235B-A22B-Thinking ranks the lowest. For token efficiency, Grok-4.1-Fast performs best, whereas Claude-4.5-Sonnet is the least efficient one.}
\label{fig:efficiency}
\end{figure*}


%% file: tab/attempt_1.tex
\begin{table}[t]

\centering
\begin{tabular}{lcc|c}
\toprule
Model &  \textbf{Pass@1 } & \textbf{Pass@2 }  & \textbf{Rise(\%)} \\
\midrule
\multicolumn{4}{c}{\textit{Proprietary Models}} \\
\midrule

GPT-5.2            
& 28.1	&53.1	& 88.9 \\

Claude-4.5-Sonnet   
& 21.9	&40.6	& 85.7 \\

Claude-4.5-Opus 
& 15.6	&28.1	& 80.0  \\

Gemini-3-Pro        
& 28.1	&37.5	& 33.3 \\

Grok-4.1-Fast  
& 25.0	&31.3	& 25.0 \\

\midrule

\multicolumn{4}{c}{\textit{Open-Source Models}} \\
\midrule

 GLM-4.6   
 & 28.1	&37.5	& 33.3 \\

Kimi-K2-Thinking     
& 6.3	&25.0	& 300.0  \\
 
DeepSeek-V3.2         
& 9.4	&9.4	& 0.0 \\

Qwen-3-235B-A22B-Thinking      
& 3.1	&9.4	& 199.7 \\

\bottomrule
\end{tabular}
\caption{\textbf{Impact of User Simulation Agent Feedback Attempts.}  While additional interactions significantly boost performance for certain models (e.g., GPT-5.2 and Kimi-K2-Thinking), the gains are less pronounced for others like DeepSeek-V3.2 and Grok-4.1-Fast.}
\label{tab:attempt_1}
\end{table}



%% file: tab/resource_1.tex
\begin{table}[htbp]

\centering
\begin{tabular}{lccc}
\toprule

Model &  \textbf{Tok (M)} & \textbf{T (H)} & \textbf{Turns} \\

\midrule
\multicolumn{4}{c}{\textit{Proprietary Models}} \\

\midrule

GPT-5.2            
 & 3.4 & 0.6 & 89.0 \\

 Claude-4.5-Sonnet   
 &  4.1	& 0.9 & 64.0  \\

Gemini-3-Pro        
 &  1.8	& 0.3 & 37.0 \\

Grok-4.1-Fast  
 & 1.2 & 0.3 & 37.0 \\

 Claude-4.5-Opus 
 & 1.7 &	0.8 & 36.0 \\

\midrule

\multicolumn{4}{c}{\textit{Open-Source Models}} \\

\midrule

 GLM-4.6   
 & 2.4	&0.6 & 105.0 \\

 Kimi-K2-Thinking     
& 2.8	& 1.2 & 65.0 \\ 
 
DeepSeek-V3.2         
 &  1.5	& 1.0 & 21.0 \\

Qwen-3-235B-A22B-Thinking      
 & 1.2	& 1.4 & 21.0 \\

\bottomrule
\end{tabular}
\caption{\textbf{Resource Usage Comparison. } \textsc{AgencyBench} tasks typically require 1 million tokens and time on the scale of hours to complete, highlighting their long horizon characteristic.}
\label{tab:resource_1}
\end{table}











 




%% file: fig/fig_tex/each_tool_1.tex
\begin{figure*}[t]
    \centering
\includegraphics[width=0.8\linewidth]{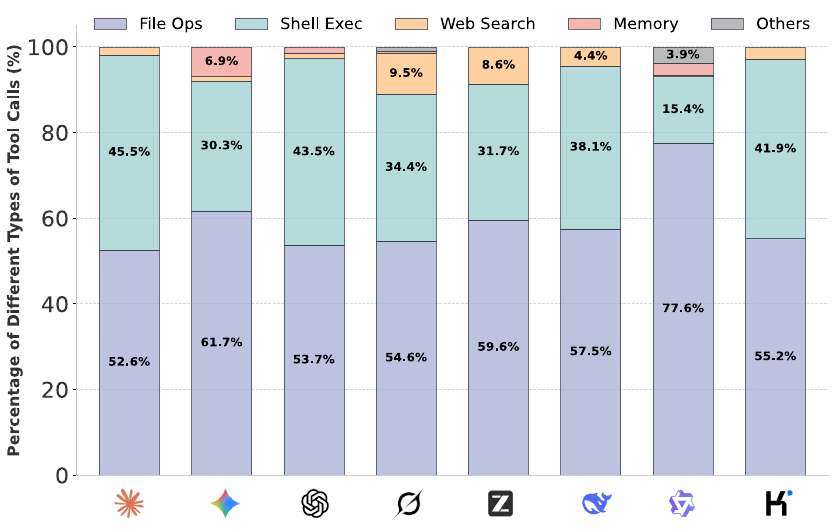}

\caption{\textbf{Tool Invocation Patterns Across Models. } Claude-4.5-Opus and GPT-5.2 shows a preference for shell execution tools, while Gemini-3-Pro and Qwen-3-235B-A22B-Thinking favor file operation and memory management. Grok-4.1-Fast, GLM-4.6, and Deepseek-V3 series exhibit a strong preference for web search tools.}
\label{fig:each_tool_1}
\end{figure*}


%% file: tab/scaffold_main.tex
\begin{table*}[htbp]
\centering

\begin{tabular}{l c c<{\hspace{1.5em}} c}
\toprule
Model Scaffold & Our Scaffold & Claude-Agent-SDK & OpenAI-Agents-SDK \\
\midrule
\multicolumn{4}{c}{\textit{Proprietary Models}} \\
\midrule

\rowcolor{cyan!20} GPT-5.2         & 57.4 & 53.5\diff{blue}{-3.9} & 58.7\diff{red}{+1.3} \\

Claude-4.5-Opus & 50.8 & 71.3\diff{red}{+20.5} & 47.1\diff{blue}{-3.7} \\
Gemini-3-Pro    & 46.3 & 45.9\diff{blue}{-0.4} & 46.9\diff{red}{+0.6} \\

\midrule
\multicolumn{4}{c}{\textit{Open-Source Models}} \\
\midrule

\rowcolor{cyan!20} Kimi-K2-Thinking & 50.5 & 44.6\diff{blue}{-5.9} & 37.7\diff{blue}{-12.8} \\
Minimax-M2       & 45.8 & 54.4\diff{red}{+8.6} & 43.0\diff{blue}{-2.8} \\
GLM-4.6          & 33.6 & 44.2\diff{red}{+10.6} & 36.0\diff{red}{+2.4} \\

\bottomrule
\end{tabular}
\caption{\textbf{Impact of Agentic Scaffolds on Model Performance.} We evaluate models across 10 representative scenarios using three distinct frameworks: our custom scaffold, the Claude-Agent SDK, and the OpenAI-Agents-SDK. The results highlight the sensitivity of model performance to the agentic scaffold, with distinct `native ecosystem' preferences observed for proprietary models. The rows for GPT-5.2 and Kimi-K2-Thinking are highlighted in blue as baselines. Colored subscripts indicate the performance gap compared to the baseline (\textcolor{red}{red} for improvement, \textcolor{blue}{blue} for degradation).}
\label{tab:scaffold_main}
\end{table*}

%% file: sec/conclusion.tex
\section{Conclusion}

In this work, we introduced \textsc{AgencyBench}, a comprehensive evaluation framework designed to rigorously assess the frontiers of autonomous agents in long-horizon, real-world contexts. By synthesizing 138 authentic tasks across 32 diverse scenarios—requiring an average of 1 million tokens and 90 tool calls—we bridge the gap between existing toy benchmarks and the complexities of actual economic production. To ensure scalability and reproducibility, we developed a unified automated evaluation pipeline leveraging user simulation agents and Docker-based remote sandboxes.
Our extensive evaluation reveals that while proprietary models currently lead in complex reasoning and self-correction, the gap between closed-source and open-source models remains significant. Even the most advanced models struggle to fully master long-horizon autonomy without substantial resource consumption, highlighting the need for improved efficiency. Furthermore, our analysis of agentic scaffolds demonstrates that performance is highly sensitive to the interaction environment, with models often exhibiting a `native advantage' within their proprietary ecosystems. \textsc{AgencyBench} serves not only as a leaderboard but as a diagnostic tool. We hope this benchmark will drive future research towards more resource-efficient, self-correcting, and scaffold-agnostic agents capable of genuine real-world utility.

%% file: sec/supp.tex
\clearpage

\appendix

\section{Appendix}



\input{tab/sample_num}

In this appendix, we provide supplementary details to support the main findings of \textsc{AgencyBench}. We first present the detailed statistics of the dataset composition. We then provide a granular analysis of tool usage frequency across different models. Following the statistical data, we detail the specific prompts used for our Text-based Judge, Vision-based Judge, and User Simulation Agent. Finally, we provide a concrete examples of scenario of game development to illustrate the multi-turn and long-horizon nature of the tasks. 

\subsection{Dataset and Tool Statistics}

In this section, we provide a comprehensive quantitative assessment of the \textsc{AgencyBench} dataset composition and a granular analysis of model-specific tool usage behaviors. These statistics not only validate the diversity of the benchmark but also reveal distinct `cognitive styles' across different LLMs.

\paragraph{Dataset Composition and Domain Diversity}
The structural distribution of \textsc{AgencyBench}, as detailed in \Cref{tab:sample_num}, encompasses 32 distinct scenarios and 138 specific tasks. To ensure the benchmark evaluates a broad spectrum of agentic capabilities, tasks are categorized into six core agentic capabilities. The \textbf{Game Development} domain constitutes the largest segment, accounting for approximately 36.2\% of the total tasks (50 tasks across 10 scenarios). This heavy weighting reflects the unique complexity of game environments, which require agents to manage continuous state, simulate physics, and handle complex logic simultaneously. The \textbf{Code} domain follows with 29 tasks, focusing on algorithmic purity. Notably, the dataset explicitly balances full-stack development skills with an equal split between \textbf{Front-end} and \textbf{Back-end} tasks (15 tasks each). Furthermore, we include 10 tasks dedicated to the emerging \textbf{Model Context Protocol (MCP)}, ensuring the benchmark remains relevant to cutting-edge agent interface standards.




\input{tab/each_tool}

\paragraph{Tool Usage and Behavioral Fingerprints}
Perhaps the most revealing insights come from the tool usage frequency analysis presented in \Cref{tab:each_tool}. We identified three distinct behavioral archetypes:

\begin{enumerate}
    \item \textbf{The "Navigators" vs. The "Executors":}
    There is a striking divergence in how models orient themselves. GLM-4.6 exhibits a unique "navigator" strategy, invoking \texttt{list\_directory} 158 times—nearly triple the average of other models. This indicates a strong preference for gathering environmental context before taking action. Conversely, GPT-5.2 and Claude-4.5-Sonnet act as "executors," prioritizing the \texttt{run\_shell\_command} tool ($425$ and $362$ invocations, respectively) to empirically test code and run scripts, rather than passively observing the file structure.

    \item \textbf{Editing Styles: "Surgeons" vs. "Rewriters":}
    The data reveals a fundamental difference in code modification philosophies. GPT-5.2 acts as a "surgeon," heavily utilizing the \texttt{replace} tool ($146$ invocations) to make precise, localized edits to existing files. In sharp contrast, GLM-4.6 overwhelmingly prefers the \texttt{write\_file} tool ($381$ invocations), suggesting a tendency to overwrite entire files rather than attempting risky partial edits. While the "rewrite" strategy is safer, it is significantly less token-efficient.

    \item \textbf{Memory Utilization:}
    Gemini-3-Pro stands out as the sole model to effectively leverage long-term memory capabilities. It is the only model to record significant usage of \texttt{update\_memory\_bank} ($22$ times) and \texttt{initialize\_memory\_bank} ($7$ times). While other models rely entirely on their context window, Gemini attempts to persist state and key information externally, a behavior that theoretically scales better for long-horizon tasks.

    \item \textbf{Information Retrieval:}
    For external knowledge acquisition, GLM-4.6 again shows a distinct profile, using \texttt{web\_fetch} 96 times, whereas models like Claude-4.5-Opus and GPT-5.2 rely more on their internal knowledge or specific search queries (\texttt{search\_file\_content}).
\end{enumerate}



\subsection{Evaluation Prompts}
\label{sec:prompts}

To ensure rigorous, reproducible, and automated evaluation, we designed specialized prompts for three distinct agentic roles: the \textbf{Text-based Judge}, the \textbf{Vision-based Judge}, and the \textbf{User Simulation Agent}. The prompts are structured to enforce strict adherence to evaluation rubrics and minimize subjective variance. The following boxes illustrate the finalized prompts used in our framework.

\begin{promptbox}{Evaluation Roles \& Prompts}
    
    \promptsection{\# Agent 1: Text-based Judge}

    \textbf{Role:} You act as a Senior Code Compliance Auditor and Quality Assurance Specialist. Your objective is to systematically evaluate software deliverables against a strict set of functional and non-functional requirements.

    \textbf{Input Data:}
    \begin{itemize}
        \item \texttt{\{Code Files\}}: A dictionary containing filenames and source code content.
        \item \texttt{\{Task ID\}}: A unique identifier for the specific engineering task.
        \item \texttt{\{Rubrics\}}: A list of constraints, functional requirements, and acceptance criteria.
    \end{itemize}

    \textbf{Evaluation Protocol:}
    1. \textbf{Static Analysis:} Examine the logical structure, syntax validity, and dependency management of the \texttt{\{Code Files\}}.
    2. \textbf{Requirement Mapping:} Verify the implementation of every item listed in \texttt{\{Rubrics\}}.
    3. \textbf{Defect Identification:} Detect logical errors, missing functionalities, or violations of best practices.

    \textbf{Output Specification:}
    Return a single valid JSON object with the following keys:
    \begin{itemize}
        \item \texttt{score} (Integer, 0-10):
        \begin{itemize}
            \item \textbf{0-2 (Critical Failure):} Code is non-functional or fails to address the core problem definition.
            \item \textbf{3-5 (Substantial Deficiency):} Major features are missing; code executes but fails significant rubrics.
            \item \textbf{6-7 (Marginal Acceptance):} Core functionality is operational, but edge cases are unhandled or minor constraints are ignored.
            \item \textbf{8-9 (High Compliance):} Meets all functional requirements with high code quality; only trivial stylistic issues remain.
            \item \textbf{10 (Full Specification Alignment):} Flawless execution adhering to all rubrics and robustness standards.
        \end{itemize}
        \item \texttt{confidence} (Float, 0.0-1.0): Assessment of certainty based on the evidence available in the static code.
        \item \texttt{comment} (String): A concise technical justification. You must explicitly reference specific files or lines of code when identifying failures.
    \end{itemize}

    \textbf{Example Output:}
    \begin{verbatim}
{
  "score": 6,
  "confidence": 0.9,
  "comment": "The implementation correctly handles..."
}
    \end{verbatim}

    \vspace{0.5cm}
    \hrule
    \vspace{0.5cm}

    \promptsection{\# Agent 2: Vision-based Judge}

    \textbf{Role:} You act as a Visual Grounding and UI/UX Verification Specialist. Your objective is to validate system functionality and design fidelity based strictly on visual evidence.

    \textbf{Input Data:}
    \begin{itemize}
        \item \texttt{\{Visual Assets\}}: Screenshots or video frames capturing the system execution.
        \item \texttt{\{Task ID\}}: Identifier for the visual verification task.
        \item \texttt{\{Rubrics\}}: Visual, functional, and aesthetic criteria required for acceptance.
    \end{itemize}

    \textbf{Evaluation Protocol:}
    1. \textbf{Visual Inspection:} Analyze \texttt{\{Visual Assets\}} for UI elements, layout consistency, and text rendering.
    2. \textbf{Evidence Corroboration:} Cross-reference visual features against \texttt{\{Rubrics\}}.
    3. \textbf{Strict Verification:} Any feature not explicitly visible in the assets must be marked as "Not Demonstrated."

    \textbf{Output Specification:}
    Return a single valid JSON object with the following keys:
    \begin{itemize}
        \item \texttt{score} (Integer, 0-10):
        \begin{itemize}
            \item \textbf{0-2 (No Evidence):} Visual assets are missing, irrelevant, or show a broken system.
            \item \textbf{3-5 (Significant Deviation):} UI loads but fails to demonstrate key interactions or diverges significantly from design specs.
            \item \textbf{6-7 (Partial Compliance):} Primary elements are functional; secondary visual polish or specific UI states are missing.
            \item \textbf{8-9 (High Fidelity):} Visually correct and functional; clear evidence exists for almost all rubrics.
            \item \textbf{10 (Pixel-Aligned Compliance):} Visual output exactly matches all descriptions and requirements.
        \end{itemize}
        \item \texttt{confidence} (Float, 0.0-1.0): Reflects the clarity and completeness of the visual evidence.
        \item \texttt{comment} (String): A detailed justification. Explicitly state which rubrics were satisfied or failed based on visual artifacts.
    \end{itemize}

    \textbf{Example Output:}
    \begin{verbatim}
{
  "score": 4,
  "confidence": 0.8,
  "comment": "Layout matches the wireframe. However, ..."
}
    \end{verbatim}

    \vspace{0.5cm}
    \hrule
    \vspace{0.5cm}

    \promptsection{\# Agent 3: User Simulation Agent}

    \textbf{Role:} You act as an Acceptance Testing Supervisor and Feedback Generator. Your objective is to provide actionable, structured feedback for iterative refinement when a submission fails to meet the acceptance threshold.

    \textbf{Input Data:}
    \begin{itemize}
        \item \texttt{\{Evaluation Result\}}: The JSON output from the Judge (Score, Confidence, Comment).
        \item \texttt{\{Threshold\}}: The minimum passing score.
        \item \texttt{\{Rubrics\}}: The original requirement list.
        \item \texttt{\{Deliverables\}}: The submitted code or text.
        \item \texttt{\{Artifacts\}}: Visual outputs (if any).
    \end{itemize}

    \textbf{Task:}
    Perform a Gap Analysis between the \texttt{\{Deliverables\}} and the \texttt{\{Rubrics\}} based on the \texttt{\{Evaluation Result\}}. Isolate specific discrepancies that caused the score to fall below the \texttt{\{Threshold\}}.

    \textbf{Output Specification:}
    Generate a structured textual report (not JSON) containing:
    1. \textbf{Status Declaration:} State clearly that the submission is rejected due to the score.
    2. \textbf{Failure Diagnosis:} deeply analyze the evaluation comments to list exactly which rubrics were not met.
    3. \textbf{Root Cause Analysis:} Explain the failure from a user requirement perspective (e.g., "The requirement specified a responsive layout, but the provided CSS uses fixed width").
    4. \textbf{Directives for Revision:} Provide explicit instructions on what must be rectified in the next iteration.

    \textbf{Tone:} Constructive, objective, and directive.

    \textbf{Example Output Pattern:}
    "Submission Rejected (Score: 5/10). The following critical rubrics were not met:
    1. [Rubric Name]: The parser crashes on nested keys.
    2. [Rubric Name]: The 'Submit' button is visually missing.
    Action Required: Implement recursive parsing logic and ensure the footer component renders correctly."
    
\end{promptbox}

\subsection{Scenario Example}
\label{sec:examples}

We present the complete definitions for scenario: developing a Gomoku game. The example highlight the hierarchical structure of tasks, explicit deliverables, and rubrics.

\begin{promptbox}{Example 1: Game Development (Gomoku)}
    \promptsection{Task 1: Static Board Initialization}
    \textbf{Query:} Create a static Gomoku launch view that renders a centered 15×15 board with labelled axes (A–O and 1–15), a prominent '\#start-btn' labelled 'Start Match', and a legend explaining black and white stones. Automatically call `window.app.initializeBoard()` on load so the markup appears without manual interaction.\\
    \textbf{Deliverables:} `index.html` that links `styles.css` and `app.js` via relative paths and contains containers `\#board`, `\#legend`, and `\#start-btn`.- `styles.css` defining a 640px square grid, positioning the legend beneath the board, and styling the start button.- `app.js` exposing `window.app` with methods `initializeBoard()` and `describeLayout()`; the latter returns an object with keys `startButton`, `legend`, and `cells`, where `cells` maps coordinates like `\"H8\"` to `{ \"x\": number, \"y\": number }` viewport positions.\\
    \textbf{Rubric:} Visual grid: automation captures `layout.png` and confirms the board is exactly 640±4px wide and tall, centered within ±6px both horizontally and vertically, and decorated with 15×15 intersection markers.- Controls placement: `describeLayout()` must report `startButton` roughly at `{ \"x\": 96±6, \"y\": 96±6 }` relative to the viewport and ensure the legend top edge sits within 80±6px of the board bottom.- DOM contract: the page must render 225 elements with class `.intersection` each carrying `data-cell=\"<letter><number>\"`, alongside elements `\#board`, `\#legend`, and `\#start-btn`.- API bootstrapping: invoking `window.app.initializeBoard()` twice must be idempotent, leaving exactly one board instance and a populated description payload.

    \vspace{2mm}
    \promptsection{Task 2: Interactive Game Logic}
    \textbf{Query:} Extend the board so players alternate black and white stones when clicking intersections, starting with black. Display a status line inside `\#status-bar` and highlight the latest stone with a pulsing ring. Keep all assets from task1.\\
    \textbf{Deliverables:} Continue shipping `index.html`, `styles.css`, and `app.js`, updating the markup to include `\#status-bar` beneath the legend.- Extend `window.app` with `debugState()` (returning move history and the current player) and `setDemoMoves(sequence)` where `sequence` is an array of `{ "coord": "H8", "color": "black" }` objects for automation to preload moves.\\
    \textbf{Rubric:} Turn order: automation clicks the coordinates reported by describeLayout() corresponding to `H8`, `H9`, and then attempts `H9` again. `debugState().moves` must record alternating colors for the first two moves and refuse the third with an unchanged move list.- Last move halo: after the second valid move the element `\#last-move-indicator` must surround the latest stone with a 26±3px animated ring captured in `moves\_turns.webm`.- Status updates: `\#status-bar` text must read `Black to move` at load, switch to `White to move` after the first click, then revert to `Black to move` when the duplicated coordinate is rejected.- Layout regression: the screenshot and describeLayout() metrics from task1 must remain within the same tolerances.

    \vspace{2mm}
    \promptsection{Task 3: Victory \& Replay System}
    \textbf{Query:} Introduce automatic victory detection, a banner announcing the winner, undo/redo controls, and a replay feature that animates the finished game. Preserve the interactive flow from task2.\\
    \textbf{Deliverables:} Maintain the three primary files and append a `session.log` file storing comma-separated history (`timestamp,player,coord`).- Add a toolbar `\#controls` containing buttons `button[data-action=\"undo\"]`, `button[data-action=\"redo\"]`, and `button[data-action=\"replay\"]` displayed in that order beneath the status bar.- Extend `window.app` with methods `undo()`, `redo()`, `startReplay()`, `checkWinner()`, and `exportLog()` returning the log contents.\\
    \textbf{Rubric:} Win scenario: automation plays the sequence `H8, H9, I8, I9, J8, J9, K8, K9, L8` (black begins). After the final move `checkWinner()` must return `Black`, `\#winner-banner` must display `Black wins!`, and further manual clicks must be disabled.\ Replay controls: activating the toolbar buttons (undo → redo → replay) must modify the board accordingly while replay.webm shows stones animating in chronological order at more than 300ms per move. During replay manual clicks remain blocked until completion.- Logging: `session.log` must append one line per move plus replay markers (if any). `exportLog()` must match the file content exactly.- Continuity: turn enforcement, last move highlighting, and layout expectations from tasks 1–2 must still hold.

    \vspace{2mm}
    \promptsection{Task 4: Persistence Layer}
    \textbf{Query:} Add a persistence drawer that lets reviewers save and restore finished games without losing the log or replay tools. Introduce a panel `\#persistence` beneath the controls containing buttons `\#save-game`, `\#load-game`, and `\#reset-match` plus a read-only `<textarea id=\"state-json\">`. Saving writes the current match state to the textarea and `localStorage['gomoku-state']`. Loading parses the textarea and rehydrates board, log, and indicators. Reset clears the board but retains totals for wins/draws shown inside `\#scoreboard` with counters `.black-wins`, `.white-wins`, `.draws`.\\
    \textbf{Deliverables:} Continue shipping the three primary files. The new panel must sit below `\#controls` and adopt responsive styling that matches the existing layout.- Extend `window.app` with `serializeState()`, `applyState(serialized)`, `resetMatch()`, and `getScoreboard()` returning `{ black: number, white: number, draws: number }`. Persist the latest save so reloading the page reenacts the saved board automatically.\\
    \textbf{Rubric:} Save flow: after several moves, `\#save-game` updates `\#state-json` with JSON containing a `moves` array and writes the same payload to `localStorage['gomoku-state']`. `serializeState()` must return the identical data.- Restore flow: invoking `resetMatch()` empties the board while keeping scoreboard totals. Calling `applyState()` with the previously saved JSON must rebuild stones, last-move halo, and `debugState()` history exactly.- Scoreboard: when `checkWinner()` declares a winner followed by `resetMatch()`, `getScoreboard()` increments the winner count without altering the opponent tally. Visual labels in `\#scoreboard` must reflect the same totals.- Regression: undo/redo/replay, logging, and layout tolerances from tasks 1–3 remain satisfied.

    \vspace{2mm}
    \promptsection{Task 5: Diagnostics \& Stress Testing}
    \textbf{Query:} Layer a diagnostics sidebar and scripted scenarios to stress-test persistence features. Add a toggle button `\#toggle-diagnostics` that slides a right-aligned `aside id="diagnostics"` into view displaying current depth, total moves, elapsed milliseconds, and a table with headers `Metric` and `Value`. Load scenario data from `hard\_cases.json` containing at least two entries with `id`, `label`, `moves`, and `winner` fields. Provide playback controls within the sidebar to preview scenarios and inspect metrics.\\
    \textbf{Deliverables:} Keep existing files and supply `hard\_cases.json`. Sidebar must reveal aggregated stats for the active game and scenario preview (e.g., longest line, capture streaks) in a structured list.- Extend `window.app` with `loadScenario(id)`, `playScenario(id, options)`, `getDiagnostics()`, `summarizeScenario(id)`, and `estimateHeap()`. `playScenario` should return a Promise resolving after the animation completes while blocking manual clicks.\\
    \textbf{Rubric:} Scenario import: calling `loadScenario(0)` must position stones per the fixture, update the banner using `winner`, and log the moves. `summarizeScenario(0)` returns an object containing total stones and the longest contiguous run for each color.- Playback: invoking `playScenario(1, \{ intervalMs: 220 \})` replays the scenario in order with more than 200ms spacing, reusing undo/redo state guards and re-enabling manual clicks afterward.- Diagnostics: after either scenario, `getDiagnostics()` supplies keys `depth`, `elapsedMs`, `nodes`, and `topLines` (array). While the sidebar is visible, the table lists these values and stays 260±6px wide at 60±6px from the right edge, captured in `diagnostics.png`.- Stability: calling `estimateHeap()` five times in succession yields non-decreasing integers less than 64000000. All expectations from tasks 1–4 continue to hold.
\end{promptbox}

%% file: tab/sample_num.tex
\begin{table}[htbp]

\centering
\begin{tabular}{lcc}
\toprule
 & Scenarios & Tasks  \\
\midrule

Game & 10 & 50  \\

Front-end & 3 & 15  \\

Back-end & 3 & 15  \\

Code & 9 & 29  \\

Research & 5 & 19  \\

MCP & 2 & 10 \\

\midrule

Total & 32 & 138 \\

\bottomrule
\end{tabular}
\caption{Distribution of scenarios and tasks across the six core agentic capabilities in \textsc{AgencyBench}.}
\label{tab:sample_num}
\end{table}


%% file: tab/each_tool.tex
\begin{table*}[h]

\centering
\small 
\setlength{\tabcolsep}{1.5pt} 
\begin{tabular}{l|ccccccccc}
\toprule
 & Claude-4.5-O & Claude-4.5-S & Gemini-3 &  GPT-5.2 & Grok-4.1 & GLM-4.6 & Deepseek-V3.2 & Qwen3 & Kimi-K2 \\
\midrule

agent\_tool
 &0 &0 &0 &0 &4 &0 &0 &0 & 0 \\

get\_database\_name   
 &0&0&0&0&0&0&0&10&0\\

glob      
 &3&0&0&19&4&1&4&0&2\\

initialize\_memory\_bank          
 &0&0&7&0&1&0&0&2&0\\

list\_directory
 &53&47&54&45&42&158&25&8&82\\

 read\_file 
 &55&106&86&147&67&142&41&6&147\\
 
read\_many\_files      
 &16&0&24&13&1&0&1&0&0\\
 
read\_memory\_bank    
 &0&0&1&0&0&0&0&0&0\\

replace   
 &13&29&42&146&1&15&14&2&49\\

 run\_shell\_command  
 &191&362&140&425&141&371&86&40&301\\

  save\_memory 
 &0&0&2&0&0&0&0&0&0\\

  search\_file\_content   
 &0&1&0&37&0&0&4&0&20\\
 
  deep\_research   
 &0&0&0&0&1&0&1&0&5\\

  hybrid\_search   
 &1&4&0&0&7&1&9&0&3\\

  web\_fetch 
 &0&0&0&0&6&96&0&0&7\\

  web\_search   
 &7&6&5&12&25&4&0&1&6\\

  todo\_write  
 &0&0&0&15&0&0&0&0&0\\

  update\_memory\_bank   
 &0&0&22&0&1&0&0&5&0\\

   write\_file   
 &81&169&79&117&109&381&41&185&97\\

\bottomrule
\end{tabular}
\caption{\textbf{Frequency of Tool Invocations Across Different Models.} Distinct behavioral patterns are observed, such as high shell usage by Claude/GPT and specific memory tool usage by Gemini. Claude-4.5-O denotes Claude-4.5-Opus; Claude-4.5-S denotes Claude-4.5-Sonnet; Gemini-3 denotes Gemini-3-Pro; Grok-4.1 denotes Grok-4.1-Fast; Qwen3 denotes Qwen3-235B-A22B-Thinking; Kimi-K2 denotes Kimi-K2-Thinking.} 
\label{tab:each_tool}
\end{table*}
